\documentclass[sigconf]{acmart}

\usepackage{graphicx}
\usepackage{amsmath}
\usepackage{booktabs}
\usepackage{multirow}
\usepackage{chngpage}
\usepackage{bbm}
\usepackage{enumitem}
\usepackage{subcaption}
\usepackage{xcolor}
\usepackage{lipsum}
\usepackage{tcolorbox}

\newcommand{\cutparagraphup}{\vspace*{0.06in}}
\newcommand{\cutcaptionup}{\vspace*{-0.03in}}

\newcommand{\model}[1]{\textsf{\small{AdStyle}}}

\renewcommand{\eqref}[1]{\ref{#1}}
\AtBeginDocument{%
  }

\setcopyright{acmlicensed}
\copyrightyear{2025}
\acmYear{2025}
\acmConference[WWW '25]{Proceedings of the ACM Web Conference 2025}{April 28-May 2, 2025}{Sydney, NSW, Australia}
\acmBooktitle{Proceedings of the ACM Web Conference 2025 (WWW '25), April 28-May 2, 2025, Sydney, NSW, Australia}
\acmDOI{10.1145/3696410.3714569}
\acmISBN{979-8-4007-1274-6/25/04}

\begin{document}

\settopmatter{authorsperrow=3, printacmref=true}

\author{Sungwon Park}
\authornote{Equal contribution to this work}
\affiliation{%
  \institution{KAIST}
  \city{Daejeon}
  \country{South Korea}
}
\email{psw0416@kaist.ac.kr}

\author{Sungwon Han}
\authornotemark[1]
\affiliation{%
  \institution{KAIST}
  \city{Daejeon}
  \country{South Korea}
}
\email{lion4151@kaist.ac.kr}

\author{Xing Xie}
\affiliation{%
  \institution{Microsoft Research Asia}
  \city{Beijing}
  \country{China}
}
\email{xingx@microsoft.com}

\author{Jae-Gil Lee}
\affiliation{%
  \institution{KAIST}
  \city{Daejeon}
  \country{South Korea}
}
\email{jaegil@kaist.ac.kr}

\author{Meeyoung Cha}
\affiliation{%
  \institution{MPI-SP \& KAIST}
  \city{Bochum}
  \country{Germany}
}

\email{meeyoungcha@kaist.ac.kr}

\renewcommand{\shortauthors}{Sungwon Park, Sungwon Han, Xing Xie, Jae-Gil Lee, and Meeyoung Cha}
\title{Adversarial Style Augmentation via Large Language Model
for Robust Fake News Detection}

\begin{abstract}
The spread of fake news harms individuals and presents a critical social challenge that must be addressed. 
Although numerous algorithmic and insightful features have been developed to detect fake news, many of these features can be manipulated with style-conversion attacks, especially with the emergence of advanced language models, making it more difficult to differentiate from genuine news.
This study proposes adversarial style augmentation, \model{}, designed to train a fake news detector that remains robust against various style-conversion attacks.
The primary mechanism involves the strategic use of LLMs to automatically generate a diverse and coherent array of style-conversion attack prompts, enhancing the generation of particularly challenging prompts for the detector. 
Experiments indicate that our augmentation strategy significantly improves robustness and detection performance when evaluated on fake news benchmark datasets.
\looseness=-1
\end{abstract}



\maketitle

\section{Introduction}
In today's digital landscape, people seek information through various channels, including social media. News content continues to hold substantial importance on these platforms, significantly shaping individuals' thoughts and decisions. However, the ease of sharing and consuming information outside traditional news outlets has introduced several challenges. Among them is the rise of alternative media and news-like content, which can contain completely or partially false information~\cite{NewsScience2018}. The spread of such misinformation has negative impacts on individuals and is considered a major social challenge that requires urgent attention~\cite{cha2021prevalence}.

Given the vast amount of information shared on social platforms, manually detecting fake news has become impractical. Consequently, these efforts are increasingly supported and managed by algorithms.
For example, sentiment and topic features, as well as the temporal and structural patterns of diffusion networks, have proven effective in detecting fake news~\cite{kownICDM2013,potthast2018stylometric,rashkin2017truth}. These patterns can be utilized by machine learning and deep learning~\cite{maIJCAI2016,shu2019defend,perez2018automatic}.
Large language models (LLMs) have also been used to learn the textual style of fake news by extracting sentence embeddings to train detectors~\cite{factualityNMI2024,hu2024bad,chen2023can}. \looseness=-1

Although certain textual styles or linguistic cues can indicate fake news, attackers can circumvent such detection by rearranging the order of subjects and objects or by imitating specific styles~\cite{koenders2021vulnerable,zhou2019fake}.
LLMs can easily paraphrase sentences in any desired manner through prompts (e.g., "change this text into a NYTimes style"), making it challenging to differentiate AI-generated news from authentic ones. These manipulations, referred to as \textsf{style-conversion attacks}, significantly complicate the task of maintaining information veracity in the digital media landscape~\cite{koenders2021vulnerable,zhou2019fake}.
\looseness=-1

In this study, we present \model{}, an adversarial style augmentation method designed to train a fake news detector that can withstand various style-conversion attacks.
Unlike previous methods that relied on predefined style-conversion prompts, our approach identifies prompts to generate adversarial style augmentations tailored to a specific detector.
This process adds noise to the style features within the detector's decision boundary, while preserving content-wise integrity.
To customize prompts that are adversarial to the detector, we implement an automated prompt engineering technique, similar to~\cite{yang2023large}.
By providing style-conversion prompts and evaluating the detector's performance under these augmentations, the LLM can discern patterns between prompts and performance, facilitating search for the most effective prompts. \looseness=-1

When evaluated on fake news benchmarks such as PolitiFact, GossipCop, and Constraint under various style-conversion attack scenarios, our augmentation strategy exhibited enhanced robustness and detection performance compared to current state-of-the-art methods.
Our strategy preserves the content of the sentence while altering its structure to increase perplexity according to the LLM-based detector. This adjustment results in a sentence structure that the LLM has not frequently encountered during its pre-training phase, effectively deterring malicious style-conversion attacks, such as paraphrasing.
Additionally, our method serves as an augmentation strategy that can be integrated with any existing detection models, regardless of their design.

We release the code and implementation details of our fake news detection against style-conversion attacks for the broader use and greater impact within the research community and industry: \url{https://github.com/deu30303/AdStyle}.\looseness=-1

\section{Related Work}

\subsection{Automated Detection of Fake News}

Among the machine-driven fake news detection methods are learning methodologies extract textual features from fake news texts using benchmark datasets and ground-truth labels~\cite{reis2019supervised}. 
These textual features can include deep features based on artificial neural networks~\cite{shu2019defend} as well as manually defined features such as sentiment and political bias~\cite{kownICDM2013,potthast2018stylometric,rashkin2017truth}. 
Other approaches include domain adaptation methods to enhance detection generalizability across various domains and topics~\cite{mosallanezhad2022domain,nan2022improving}, and knowledge-based methods that rely on external data to distinguish false information~\cite{dun2021kan}.
With the advent of LLMs, new methods have emerged that utilize LLMs' prior knowledge to identify fake news~\cite{hu2024bad}, including research on detecting machine-generated fake news~\cite{chen2023can}. One approach involves using LLMs to generate synthetic reactions and comments on news articles from diverse perspectives, which are then leveraged for fake news detection~\cite{wan2024dell}. Furthermore, LLM-generated logic has been employed to enhance the interpretability of fake news detection models~\cite{liu2024teller}. Despite these advancements, fake news detection using LLMs remains a challenging task due to factuality hallucinations, which can undermine reliability~\cite{augenstein2024factuality}.\looseness=-1

\subsection{Attack on Fake News Detection}
Various attack methods have been studied to test the robustness of fake news detection~\cite{koenders2021vulnerable,zhou2019fake}. 
These methods include injecting misinformation by rearranging subjects and objects or causes and effects while retaining the textual features used for detection~\cite{koenders2021vulnerable}. 
Other techniques involve creating fact distortions by modifying or exaggerating words related to people, time, or places, while preserving sentence structure~\cite{zhou2019fake}.  Another line of research simulates the behavior of malicious users on social media by manipulating news article comments or employing a multi-agent reinforcement learning framework to generate adversarial content~\cite{le2020malcom,wang2023attacking}. Recently, methods leveraging the text generation capabilities of LLMs to modify the writing style of sentences have also been proposed~\cite{wu2023fake}. In this study, we propose an adversarial training method to mitigate style-conversion attacks in fake news detection. Our approach utilizes automated prompt engineering to identify the most effective style-conversion prompts for a given fake news dataset~\cite{yang2023large,zhou2022large}. \looseness=-1

\begin{figure*}[t!]
\centering
\includegraphics[width=0.9\textwidth]{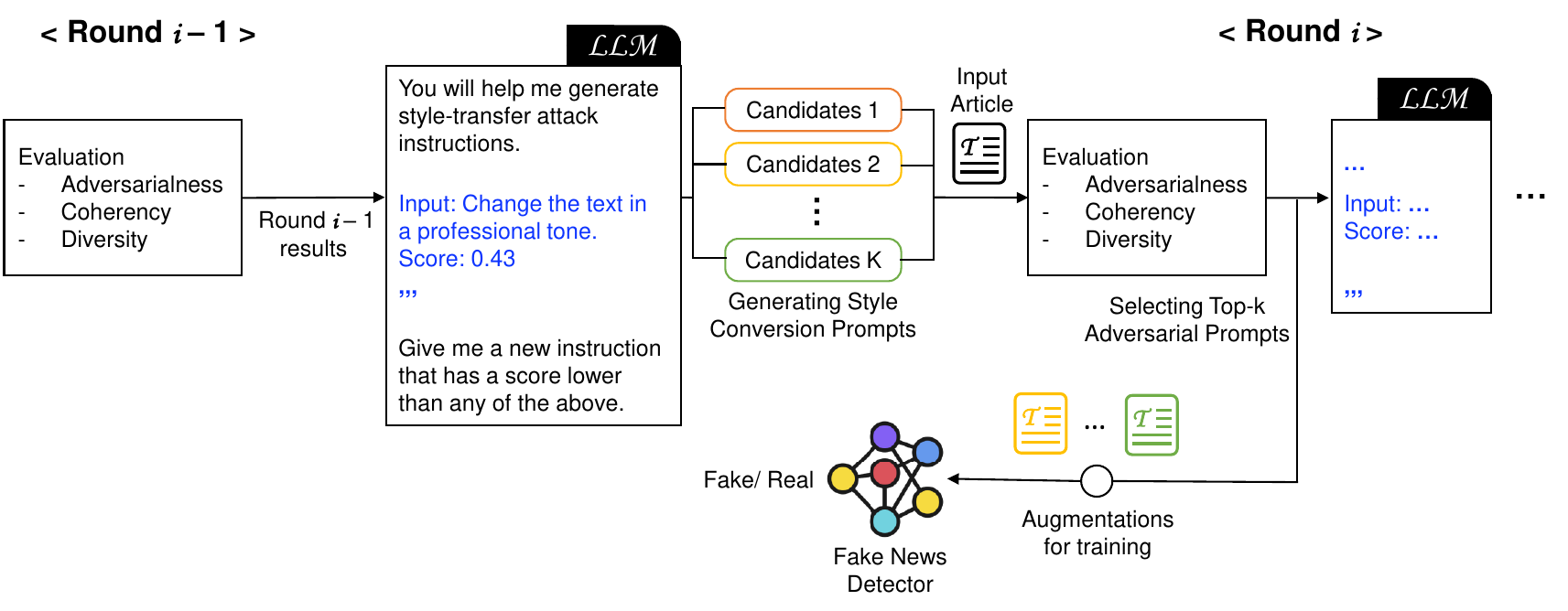}
\cutcaptionup
\caption{Illustration of \model{}. The model undergoes multiple rounds of training. In each round, it utilizes the style-conversion prompt and prediction confusion score from the previous round to create prompt candidates aimed at maximizing the detector model's confusion. Subsequently, a subset of the training dataset is used to select the top-$k$ prompts, based on factors such as adversarialness, coherency, and diversity. These selected prompts are then employed as augmentations in the training process. Repetition of this process allows the model to consider a wide range of adversarial prompts, making the director more resilient to future attacks.}
\label{fig:main_model}
\end{figure*}

\section{Method}
\subsection{Overview}
Let $\mathcal{D} = \{(\mathbf{d}_i, y_i) \}_{i=1}^N$ be a dataset containing news $\mathbf{d}_i$ and the corresponding ground-truth binary veracity label $y_i$ (indicating whether the news is true or fake). 
Each news item $\mathbf{d}_i$ is composed of natural language-based text.
Our goal is to train a fake news detector $f$ based on the labeled dataset and predict veracity labels. 
In doing so, we ensure that a detector remains robust even when an attacker perturbs the textual style, such as the order and format, while preserving the meaning of the sentences.  \looseness=-1

Figure~\ref{fig:main_model} illustrates the workflow. 
Our method, called \model{}, generates style-conversion prompts and performs augmentation over multiple rounds of the following process. 
Firstly, leveraging the reasoning ability of LLM, adversarial style-conversion prompt candidates are generated. 
These style-conversion prompts contain instructions on how to transform the given sentences and are used as inputs to the LLM along with the original sentences for conversion. 
Following the automated prompt engineering~\cite{yang2023large}, the style-conversion prompts and detector prediction score pairs used in previous rounds are included as in-context demonstrations to guide the LLM in searching prompt candidates that maximally confuse the detector (Section~\ref{sec:3.2}).
Next, a subset of the dataset is selected, and these discovered candidates are applied to perform conversions. 
The converted samples are then evaluated to determine how much they confuse the detector. 
From these candidates, the top-$k$ prompts that are diverse and maintain the original content's meaning while most effectively confusing the detector are selected (Section~\ref{sec:3.3}). 
These selected prompts are used as an augmentation method to train the detection model in the current round.
We describe each step's details below. \looseness=-1

\begin{figure}[t!]
\begin{tcolorbox}[
  colback=black!0!white, colframe=black!20!white, colbacktitle=black!10!white, coltitle=blue!20!black ]

\sffamily
\small{

Now you will help me minimize a fake news detector performance value with style transfer attack instructions. I have a set of style transfer attack instructions that I will use to minimize the fake news detector's performance. 
The pairs are arranged in descending order based on their function values, where lower values are better. \\
\\
\textcolor{blue}{Input: [Style-Conversion Prompt Example]} \\
\textcolor{blue}{Score: [Prediction Confusion Score]} \\
\\
\textcolor{blue}{Input: ...}
\textcolor{blue}{...} \\
\\
Give me a new style transfer attack instruction that is different from all pairs above and has a lower performance value.
}
\end{tcolorbox}
\cutcaptionup
\captionof{figure}{Prompt for generating adversarial style-conversion prompts. The text in blue represents the score trajectory, and the remaining text represents the problem description. \looseness=-1}
\label{fig:prompt_for_generation}            
\end{figure}

\subsection{Adversarial Style-Conversion Prompts}
\label{sec:3.2}
The style-conversion prompts we generate are instructions that perturb only the textual style, such as the structure or format of the sentence, while preserving the content of the given input text. 
For example, an instruction like ``Rewrite the following article in an objective and professional tone'' can transform the stylistic features of fake news to resemble authentic news.
However, if these instructions are heuristically defined, they may not align with the detector's actual decision boundary, thereby reducing training efficiency. 
Furthermore, if the instructions remain fixed throughout the training process, the detector may memorize these conversion patterns, leading to overfitting.

Instead of predefining and fixing the instructions for conversion, we identify adversarial conversion prompts that introduce noise directed toward the decision boundary of the current detector, thereby maximizing prediction confusion. 
This approach is similar to adversarial training commonly used in the computer vision domain~\cite{madry2018towards}, which enhances robustness against slight input perturbations.
However, for textual data, the discrete nature of the input makes it challenging to add noise directly through the detector's gradient as in the image domain. 
To address this, we measure prediction confusion using conversion prompts (i.e., prediction confusion score) and feed this prompt-score pair to the LLM for generating adversarial prompts, inspired by automated prompt engineering techniques. \looseness=-1

Figure~\ref{fig:prompt_for_generation} shows an example of the LLM input used to generate adversarial style-conversion prompts. 
The LLM input consists of the problem description and the score trajectory components. We describe each component in turn.\looseness=-1

\cutparagraphup
\noindent\textbf{Problem description component.}
This component includes the problem description, the objective, and constraints on the response necessary for generating style-conversion prompts. 
For example, a sentence like ``Minimize a fake news detector performance value with style transfer attack instruction'' informs the LLM of the intent behind the conversion prompt.  

\cutparagraphup
\noindent\textbf{Score trajectory component.}
Previous research has demonstrated that LLMs can learn patterns from in-context demonstrations provided as input~\cite{dong2022survey,yang2023large}. 
The score trajectory component leverages this ability by providing style-conversion prompts from previous rounds and their corresponding prediction confusion scores in the form of in-context demonstrations. In the first round, a predefined set of prompts is excerpted from~\cite{wu2023fake}.
The score is calculated by selecting a subset $\mathcal{B}$ from the entire training dataset $\mathcal{D} = \{(\mathbf{d}_i, y_i) \}_{i=1}^N$, applying a conversion prompt $c$ to create a new set $\mathcal{B}^c = \{(\mathbf{d}^c_i, y_i)  \}_{i=1}^M$, where $\mathbf{d}^c_i = \text{Convert}(c, \mathbf{d}_i), N>>M$, and then measuring the AUC score between the predictions and the ground-truth labels when $\mathcal{B}^c$ is fed into the detector. Specifically, the score of a conversion prompt $s_c$ is defined as: \looseness=-1
\begin{align}
       s_c = | 0.5 - \text{AUC}( \{y_i \}_{i=1}^M,   \{ f(\mathbf{d}^c_i)   \}_{i=1}^M  ) |.
       \label{eq:confusion_score}
\end{align}

A lower score indicates a higher level of prediction confusion, implying that the conversion prompt has caused the detector's predictions to become random with respect to the labels. 
Finding conversion prompts that lead to high confusion (i.e., low score) suggests that the detector has not yet learned to handle those stylistic features, and the conversion prompts have placed the samples near the detector's decision boundary, making them difficult to distinguish. 
By providing these in-context demonstrations, the LLM can generate conversion prompts that differ from previous ones while maximizing confusion (i.e., minimizing the score). 
It is important to avoid selecting conversion prompts that flip the detector’s original predictions (i.e., AUC $<<$ 0.5), as such cases will disrupt the content and stylistic features.
We extract $S$ style-conversion prompts at a time using the input described above.

\subsection{Selecting Top-$k$ Adversarial Prompts}
\label{sec:3.3}
Using all the style-conversion prompt candidates generated by the LLM can be computationally intensive, and not all prompts may be suitable for augmentation. 
For example, the set of style-conversion prompts used for augmentation should each provide adversarial perturbations that confuse the detector (i.e., adversarialness). 
While altering the textual style, the prompts should not change the meaning of the sentences to prevent label noise (i.e., coherency). 
The more diverse the set of conversion directions covered by the prompts, the more efficient the augmentation process (i.e., diversity).

To select a set of style-conversion prompts that satisfies these three criteria, we propose a selection strategy. 
Given a conversion prompt $c$, we first extract embedding vectors for the input texts in both the training subset $\mathcal{B}$ and the subset $\mathcal{B}^c$ converted by $c$ using a large language model $g$ such as BERT. 
We then compute the average embedding vectors for each subset and calculate the vector difference $\mathbf{z}_c$. \looseness=-1
\begin{align}
\mathbf{z} &= \frac{1}{|\mathcal{B}|} \sum_{\mathbf{d}_i \in \mathcal{B}}{g(\mathbf{d}_i)},\ \ \  
\mathbf{z}’ = \frac{1}{|\mathcal{B}’|} \sum_{\mathbf{d}^c_i \in \mathcal{B}’}{g(\mathbf{d}^c_i)} \nonumber \\
\mathbf{z}_c &= \mathbf{z}’ - \mathbf{z} \label{eq:z_c} 
\end{align}
Here, $\mathbf{z}_c$ represents the average change in embedding direction due to the conversion prompt $c$. 
We then calculate the adversarialness scale $s^c_{\text{adv}}$ and coherency scale $s^c_{\text{coh}}$ for each conversion prompt $c$, and rescale $\mathbf{z}_c$ accordingly (i.e.,  $\mathbf{\hat{z}}_c = \mathbf{z}_c \times s^c_{\text{adv}} \times s^c_{\text{coh}}$). 
Finally, we use the $k$-means++ initialization method~\cite{arthur2007k} on these rescaled vectors to select $k$ prompts. 
The $k$-means++ initialization method helps select a diverse set of prompts that are adversarial and coherent, by choosing samples that are as far apart as possible~\cite{ash2019deep}.
The details of each scale are described below.

\cutparagraphup
\noindent\textbf{Adversarialness scale} ($s^c_{\text{adv}}$).
To measure how adversarial a given style-conversion prompt is to the detector, we newly define the
\textsf{adversarialness} scale similar to the confusion score defined in the previous section. Given the converted batch $\mathcal{B}^c$ by conversion prompt $c$, the adversarialness scale $s^c_{\text{adv}}$ is defined as:
\begin{align}
s^c_{\text{adv}} = -1.8 \cdot | \text{AUC}( \{y_i \}_{i=1}^M,   \{ f(\mathbf{d}^c_i) \}_{i=1}^M  ) - 0.5| + 1. 
\looseness=-1
\end{align}
This value increases as the AUC approaches 0.5, indicating that the prediction is more random. The coefficient 1.8 ensures that the scale ranges between 0.1 and 1, preventing it from being zero.

\cutparagraphup
\noindent\textbf{Coherency scale} ($s^c_{\text{coh}}$).
To verify that the converted text retains the same content as the original text, we check the similarity in meaning between the text pairs using an LLM. 
We create sample pairs from $\mathcal{B}$ and the converted subset $\mathcal{B}^c$, and inquire the LLM about the percentage of pairs that it considers to have the same meaning.
This percentage is used as the coherency scale $s^c_{\text{coh}}$. 
Like the adversarialness scale, this value is rescaled to range between 0.1 and 1. \looseness=-1

Finally, the selected style-conversion prompts via our score and $k$-means++ initialization method are applied to the input texts of the entire dataset $\mathcal{D}$ to create augmented samples. 
These augmented samples are then used alongside the original samples to train the detector $f$. 
The detector is trained using binary cross-entropy loss.
These prompt generation and selection processes are repeated over multiple training rounds.

\section{Experiment}
We evaluate the robustness of \model{} under diverse style-conversion attacks across multiple datasets, comparing it with contemporary baselines. We then analyze the impact of various model components on overall performance. A comprehensive evaluation is also performed on a broader range of paraphrasing attacks, including comparisons with LLM-based zero-shot and in-context learning baselines. Lastly, a qualitative analysis is conducted to explore the characteristics of the generated style-conversion prompts.
\looseness=-1

\subsection{Performance Evaluation}
\noindent\textbf{Dataset.} We use three real-world fake news benchmark datasets: (1) PolitiFact and (2) Gossipcop, released by FakeNewsNet benchmark~\cite{shu2020fakenewsnet}, which focus on political claims and celebrity rumors, respectively, and (3) Constraint~\cite{felber2021constraint}, a dataset of social media posts on COVID-19. 
Datasets were split into an 80\% training set and a 20\% test set, as described in Table~\ref{tab:ds-stats}. 

\begin{table}[!h]
\setlength{\tabcolsep}{3.5pt}
\centering
\caption{
Statistics of fake news datasets.}
\cutcaptionup
\scalebox{1.0}{
 \begin{tabular}{lccc} \toprule
 \textbf{Dataset} &  \textbf{PolitiFact} & \textbf{GossipCop} & \textbf{Constraint} \\ 
 \toprule
 \# of News Articles & 774 & 7,916 & 8,418\\
 \# of Real News & 399 & 3,958 & 4,406 \\
 \# of Fake News & 375 & 3,958 & 4,012 \\  
 \bottomrule
\end{tabular} 
}
 \label{tab:ds-stats}
\end{table}

\begin{table*}[t!]
\caption{Performance comparison with \model{} in two different scenarios—Attack, where style-conversion attacks are performed, and Clean, where no attack is performed—across three fake news datasets. For the attack scenario, we report the average AUC of four style-conversion attacks. The best results are marked in bold. The values 0.1, 0.25, and 1 indicate the proportion of the dataset used for training relative to the full dataset.  Our model demonstrates a significant performance improvement over all text-based fake news detectors in both style-conversion attack and clean scenarios. \looseness=-1}
\cutcaptionup
\centering
\scalebox{1.08}{
\begin{tabular}{l|ccc|ccc|ccc}
\toprule
\multirow{2}{*}{Attack } &  & Politifact  & &  &  Gossipcop& &  &Constraint & \\
 & 0.1 & 0.25 & 1 & 0.1 & 0.25 & 1 & 0.1 & 0.25 & 1 \\
\midrule
Vanilla & 0.6114 & 0.6904 & 0.8548 & 0.6920 & 0.7876 & 0.8453 & 0.8185 & 0.8674 & 0.8741 \\

UDA & 0.6241 & 0.7696 & 0.8564 & 0.7381 & 0.7865 & 0.8591 & 0.8615 & 0.9028 & 0.9297 \\
RADAR & 0.7218 & 0.7399 & 0.8571 & 0.7583 & 0.8170 & 0.8616 & 0.8535 & 0.8047 & 0.9086 \\
ENDEF & 0.6376 & 0.7579 & 0.8134 & 0.7405 & 0.7870 & 0.8615 & 0.8234 & 0.8950 & 0.8835 \\
SheepDog & 0.6525 & 0.8234 & 0.9009 & 0.7498 & 0.8357 & 0.8669 & 0.8926 & 0.9188 & 0.9630 \\\midrule
\model{} & \textbf{0.7833} & \textbf{0.8919} & \textbf{0.9399} & \textbf{0.8134} & \textbf{0.8389} & \textbf{0.8721} & \textbf{0.9224} & \textbf{0.9531} & \textbf{0.9716} \\ 
\bottomrule
\end{tabular}
}

\vspace{2mm}

\scalebox{1.08}{
\begin{tabular}{l|ccc|ccc|ccc}
\toprule
\multirow{2}{*}{Clean} &  & Politifact & &  &  Gossipcop& &  & Constraint & \\
 & 0.1 & 0.25 & 1 & 0.1 & 0.25 & 1 & 0.1 & 0.25 & 1 \\
\midrule
Vanilla & 0.7393 & 0.8397 & 0.9355 & 0.7096 & 0.8104 & 0.8645 & 0.9311 & 0.9682 & 0.9892 \\

UDA & 0.7404 & 0.8783 & 0.9422 & 0.7422 & 0.8022 & 0.8666 & 0.9365 & 0.9724 & \textbf{0.9899} \\
RADAR & 0.7607 & 0.8495 & 0.9314 & 0.7593 & 0.8170 & 0.8630 & 0.9446 & 0.9773 & 0.9817 \\
ENDEF & 0.7776 & 0.8823 & 0.9294 & 0.7592 & 0.7991 & 0.8738 & 0.9234 & 0.9556 & 0.9871 \\
SheepDog & 0.7248 & 0.8229 & 0.9394 & 0.7490 & 0.8411 & 0.8641 & 0.9144 & 0.9459 & 0.9785 \\\midrule
\model{} & \textbf{0.8996} & \textbf{0.9280} & \textbf{0.9460} & \textbf{0.8251} & \textbf{0.8493} & \textbf{0.8797} & \textbf{0.9509} & \textbf{0.9849} & 0.9889 \\
\bottomrule
\end{tabular}
}
\label{tab:style_attack_main_result}
\end{table*}

\noindent\textbf{Attack settings.} 
To assess robustness against style conversion attacks, we employ LLM-empowered techniques to reframe the test set using a variety of style conversion prompts, as illustrated in Figure~\ref{fig:stlye_attack}. 
Following the original literature~\cite{wu2023fake}, we use four prominent daily news sources as [publisher name]: \textbf{CNN}, \textbf{The New York Times}, \textbf{The Sun}, and \textbf{National Enquirer}.
CNN and The New York Times were chosen as representative news outlets recognized for their reputable journalism, while The Sun and National Enquirer are characterized by their tabloid style. 
We utilize OpenAI's GPT-3.5-Turbo model for reframing input sentences, with the temperature set to 0 and the top-p value set to 1 by default. 
Experiments are conducted using 10\%, 25\%, and 100\% of the complete dataset to observe the impact of augmentation on training performance, measured by AUC, across various dataset sizes.
\looseness=-1

\begin{figure}[h!]
\begin{tcolorbox}[
  colback=black!0!white, colframe=black!20!white, colbacktitle=black!10!white, coltitle=blue!20!black ]
  \sffamily
  Rewrite the following article using the style of [publisher
name]: [news article]
\end{tcolorbox}
\cutcaptionup
\captionof{figure}{Prompt for style conversion. The ``publisher name" part will be filled with the name of a representative publisher (e.g., newspaper or journal), and the ``news article" part will contain the original news text. \looseness=-1}
\label{fig:stlye_attack} 
\end{figure}

\begin{table*}[t!]
\centering
\caption{Performance comparison of ablations on the Politifact dataset. The results show the impact of style-conversion attacks using four different publishers (i.e., CNN, The New York Times, The Sun, and National Enquirer) and a clean scenario (i.e., Clean) where no attack is performed. Any modification or removal of model components leads to decreased performance. \looseness=-1}
\cutcaptionup
\scalebox{1.1}{
\begin{tabular}{l|ccccc}
\toprule
Model & CNN & The New York Times & The Sun & National Enquirer & Clean \\
\midrule
Vanilla & 0.8127 & 0.8687 & 0.8789 & 0.8591 & 0.9355 \\
Random Selection & 0.9075 & 0.9409 & 0.9355 & 0.9365 & 0.9412 \\
Class Prompt & 0.9131 & 0.9272 & 0.9333 & 0.9313 & 0.9471 \\
Adversarial only Selection & 0.9035 & 0.9343 & 0.9318 & 0.9288 & 0.9405 \\
w/o Adversarialness & 0.9039 & 0.9333 & 0.9320 & 0.9402 & \textbf{0.9473} \\
w/o Coherency & 0.9193 & 0.9315 & 0.9297 & 0.9430 & 0.9409 \\
w/o Score  trajectory & 0.8199 & 0.8917 & 0.9124 & 0.9066 & 0.9372 \\ \midrule
Full Components & \textbf{0.9174} & \textbf{0.9444} & \textbf{0.9520} & \textbf{0.9460} & 0.9460 \\
\bottomrule
\end{tabular}}
\label{tab:ablation_result}
\end{table*}

\cutparagraphup
\noindent\textbf{Baselines.} 
We implemented several existing text-based fake news detection strategies as baselines: (1) \textsf{Vanilla}, a text-based fake news detector using conventional binary cross entropy objective; 
(2) \textsf{UDA}, introduces consistency regularization objective between original text and diverse augmented variations.~\cite{xie2020unsupervised}; 
(3) \textsf{RADAR}, utilizes an adversarially trained paraphraser to generate augmented version of input sentences~\cite{hu2023radar}; 
(4) \textsf{ENDEF}, mitigates entity bias in fake news data through causal learning~\cite{zhu2022generalizing};
(5) \textsf{SheepDog}, introduces predefined style-conversion prompts to augment the styles of input text via LLM~\cite{wu2023fake}. 
We follow the original paper's setting and details for baseline implementations. 

\cutparagraphup
\noindent\textbf{Implementation details.}  
All models are evaluated under uniform experimental conditions to ensure fair comparison. 
This consistency extends to the choice of backbone network, optimizer, and learning rate. 
We utilize OpenAI's GPT-3.5-Turbo model for reframing input sentences and measuring coherency, with the temperature set to 0 and the top-p value set to 1 by default.
The training process for the detector is conducted over 10 rounds, with one training epoch per round. When evaluating the style-conversion prompts, we randomly selected 30 samples from the training dataset to apply augmentation (i.e., $M=30$). 
In each round, 30 prompt candidates were generated by the LLM (i.e., $S=30$), from which 3 were chosen for augmentation (i.e., $k=3$) by our selection strategy. 
The training utilizes the AdamW optimizer with a learning rate of 1e-5 and a batch size of 8. For measuring diversity, the text embeddings are generated using a pre-trained BERT-based uncased model from HuggingFace Transformers. 
Two V100 GPUs were utilized for all experiments.

\begin{figure}[t!]
\centering
\includegraphics[width=0.41\textwidth]{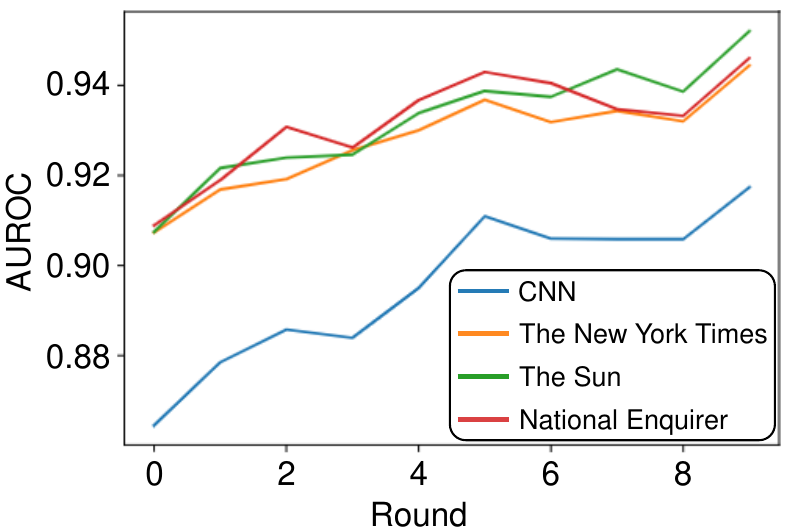} 
\cutcaptionup
\caption{
Performance changes across rounds on the PolitiFact dataset for four different style-conversion attacks. The x-axis represents the training rounds, and the y-axis represents the detector's AUC. For all attacks, the detector's performance improved as the rounds progressed. }
\label{fig:round_performance}
\end{figure}

\cutparagraphup
\noindent\textbf{Result.} Table \ref{tab:style_attack_main_result} shows results comparing the performance of detector algorithms in both clean scenarios, where no attack is performed, and adversarial scenarios, where style-conversion attacks are applied.
Due to space limitations, the average AUC results under the four different style-conversion attacks are reported. %  and detailed results are given in the Appendix (See Table~\ref{tab:style_attack_full_result})
We find that \model{} consistently outperforms the baselines, including both clean and adversarial scenarios. 
This demonstrates that our augmentation strategy enhances both the robustness and generalizability of the detector. 
The model is particularly effective when compared to other baselines and for smaller datasets.
Figure \ref{fig:round_performance} shows the AUC for each style-conversion attack over different rounds. 
The performance gradually improves as the rounds progress, indicating that continually discovering adversarial augmentations is beneficial. \looseness=-1

\subsection{Component Analysis}
We now examine the contribution of each component on our adversarial style-conversion prompts. The proposed method integrates two main modules: adversarial style-conversion prompts generation and selection. 
To evaluate their individual contributions, we conduct experiments where we either remove each component or substituted it with an alternative within the full model. 
This results in six distinct configurations for analysis: 
(1) \textbf{Full Components}: Our complete method with all components;
(2) \textbf{Random Selection}: The method that randomly select conversion prompts from candidates instead of using our selection strategy (Sec.~\ref{sec:3.3});
(3) \textbf{Class Prompt}: The method categorizes the confusion scores of conversion prompts into three levels: high, medium, or low. These categorized labels are then used as in-context demonstrations for generating adversarial style-conversion prompts (Sec.~\ref{sec:3.2}), instead of relying on continuous confusion scores $s_c$ (\ref{eq:confusion_score});
(4) \textbf{Adversarial only Selection}: The method that selects top-$k$ adversarial prompts, not considering diversity and coherence criteria;
(5) \textbf{w/o Adversarialness}: The method that omits the adversarialness criterion in our selection strategy;
(6) \textbf{w/o Coherency}: The method that omits the coherency criterion in our selection strategy;
(7) \textbf{w/o Score trajectory}: The method without score trajectory component for the style-conversion prompt generation (Sec.~\ref{sec:3.2}). 

\begin{figure}[!t]
\centering
\includegraphics[width=0.41\textwidth]{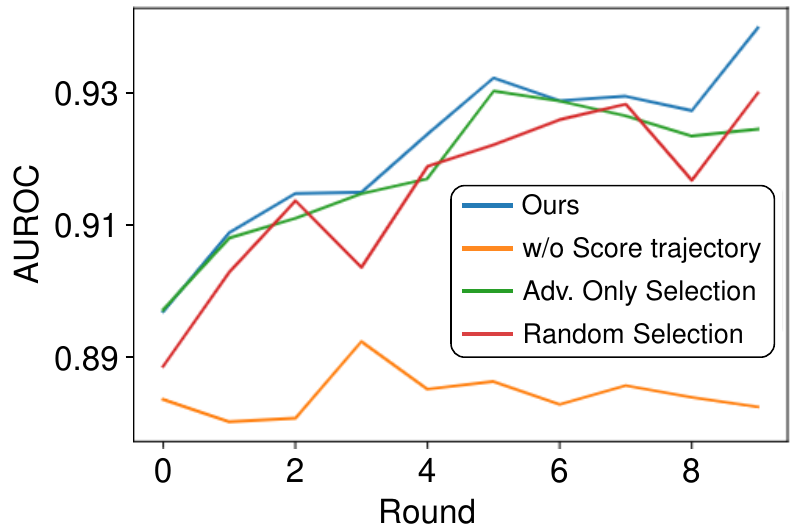} 
\cutcaptionup
\caption{Performance of selection strategies over training rounds on PolitiFact. The x-axis represents the training rounds, while the y-axis represents the detector's AUC.  \looseness=-1}
\label{fig:round_performance_select}
\end{figure}

Table \ref{tab:ablation_result} demonstrates that omitting any component leads to a decrease in performance for certain style conversions attack.  
Notably, excluding the score trajectory component when selecting the style-conversion prompt proved to be the most detrimental to performance. 
This finding suggests that the LLM can effectively identify conversion prompts that may confuse the detector through the style-conversion prompt and confusion score pairs as in-context demonstrations. 
Moreover, selectively choosing conversion prompts identified by the LLM based on specific criteria resulted in further performance improvements.
Our sampling strategy also facilitates faster convergence compared to alternative sampling methods (See Figure~\ref{fig:round_performance_select}).

\begin{table}[t!]
\caption{
Performance comparison of different LLM-based baselines on the PolitiFact Dataset. (NY: The New York Times, TS: The Sun, NE: National Enquirer) }
\cutcaptionup
\scalebox{0.98}{
\begin{tabular}{l|ccccc}
\toprule
Model  & CNN & NY & TS & NE & Clean \\ \midrule
GPT-3.5  & 0.5820 & 0.6242 & 0.6173 & 0.5274 & 0.7037 \\
GPT-4 & 0.7133 & 0.6748 & 0.7025 & 0.6021 & 0.7823 \\\bottomrule
% GPT-3.5 in-context & 0.6954 & 0.6504 & 0.6875 & 0.5754 & 0.7383 \\ \midrule
% Ours & 0.9174 & 0.9444 & 0.9520 & 0.9460 & 0.9460 \\ \bottomrule
\end{tabular}
}
 \label{tab:llm-result}
\end{table}

\subsection{Performance Analysis}
%We have demonstrated \model{}'s effectiveness. 
We here conduct analysis on how \model{} demonstrates robust and high performance across various scenarios and how it effectively enhances the detector's performance.

\cutparagraphup
\noindent\textbf{Comparison with LLM-based baselines.} 
\model{} enhanced the detector's performance by leveraging the reasoning abilities of advanced large language models like GPT-3.5. To determine whether the capabilities of an advanced large language model alone are sufficient for the fake news detection task, we compared \model{} with other LLM-based baselines: zero-shot and in-context learning-based inference with GPT-3.5. In the case of the in-context learning baseline, one example each of fake news and real news was provided as in-context demonstrations. The example instruction prompt for the LLM-based baselines are as follows:

Table~\ref{tab:llm-result} presents the comparison results on the PolitiFact dataset. Using LLMs with prompting alone (Figure~\ref{fig:llm_base_prompt}) makes it challenging to accurately determine the authenticity of news. 
Instead of `directly' leveraging the LLM's text generation capability for inference, it is more effective to use it as an augmentation tool to provide additional training signals, as demonstrated by our model.

\begin{figure}[h!]
\begin{tcolorbox}[
  colback=black!0!white, colframe=black!20!white, colbacktitle=black!10!white, coltitle=blue!20!black ]

  \sffamily
Does the following contain real or fake news? Answer in one word with either ‘Real’or ‘Fake': [news article]
\end{tcolorbox}
\cutcaptionup
\captionof{figure}{Instruction prompt for LLM-based baselines. The "news article" section contains the original news text. \looseness=-1}
\label{fig:llm_base_prompt} 
\end{figure}

\begin{table}[t!]
\caption{
Comparison under attack scenarios with Gemini-Pro on the PolitiFact dataset. \looseness=-1 } %  (NY: The New York Times, TS: The Sun, NE: National Enquirer)
\cutcaptionup
\scalebox{1.0}{
\begin{tabular}{l|cccc|c}
\toprule
\centering
Model  & CNN & NY & TS & NE  & Average \\ \midrule
Vanilla & 0.7913 & 0.8039 & 0.8882 & 0.8397 & 0.8308 \\
UDA & 0.7216 & 0.7863 & 0.8657 & 0.8119 & 0.7964\\
RADAR & 0.8023 & 0.7805 & 0.8764 & 0.8423 & 0.8254\\
ENDEF & 0.7730 & 0.7537 & 0.8439 & 0.8058 & 0.7941\\
SheepDog & 0.8487 & 0.8926 & 0.9174 & 0.8998 & 0.8896 \\\midrule
Ours & 0.8821 & 0.9120 & 0.9295 & 0.9241 & 0.9120\\ \bottomrule
\end{tabular}
}
 \label{tab:llm-result_gemini}
\end{table}

\cutparagraphup
\noindent\textbf{Robustness against a other LLM backbones.}
We assessed the robustness by examining its performance against style conversion attacks when the attacker utilizes a different LLM backbone, such as Gemini-Pro~\cite{reid2024gemini}. 
As shown in Table~\ref{tab:llm-result_gemini}, \model{} consistently outperforms the baselines even with a different LLM backbone, suggesting that the proposed approach is not overly reliant on recognizing the specific style of content generated by the backbone.

\cutparagraphup
\noindent\textbf{Robustness against other possible attack scenarios.} To validate the model's robustness against various attacks, we have conducted  comparison experiments with diverse attack scenarios to deceive the detector by altering textual style of inputs:
(1) \textsf{Adversarial prompt}: Given a news article and its label, the prompt instructs the LLM to rewrite the article to evade detection as the given label.
(2) \textsf{Summarization prompt}: A prompt instructing the LLM to summarize the news article without incorporating stylistic guidance.
(3) \textsf{In-Context prompt}: A prompt providing an example of a recent, real CNN article, instructing the LLM to rewrite the given article in the same style.
(4) \textsf{Adversarial Paraphraser}: The attack utilizes a paraphraser adversarially trained on the training dataset.
The example prompt for each attack is described in Figures \ref{fig:adv_prompt} to \ref{fig:in_context_prompt} in Appendix. 
Based on the results in Table \ref{tab:div_attack}, we confirm that our proposed model still demonstrates a performance improvement compared to other baselines against these attacks. \looseness=-1

\begin{table}[!t]
\centering
\caption{Comparison under attack scenarios on the PolitiFact Dataset (A: Adversarial prompt, B: Summarization prompt, C:  In-Context prompt D:  Adversarial Paraphraser).  }
\cutcaptionup
\scalebox{1.0}{
\begin{tabular}{l|cccc}
\toprule
Model & A & B & C & D \\ \midrule
Vanilla & 0.8881 & 0.8522 & 0.8896 & 0.8305 \\
UDA & 0.8624 & 0.8363 & 0.8924 & 0.8682 \\
RADAR & 0.9096 & 0.8754 & 0.9234 & 0.9007 \\
ENDEF & 0.8628 & 0.7978 & 0.9009 & 0.8682 \\
SheepDog & 0.9297 & 0.9205 & 0.9276 & 0.8995 \\ \midrule
Ours & 0.9456 & 0.9416 & 0.9425 & 0.9212 \\ \bottomrule
\end{tabular} 
}
\label{tab:div_attack}
\end{table}

\begin{figure}[t!]
\centering

\begin{subfigure}[t!]{0.152\textwidth}
\captionsetup{width=1.0\linewidth}
\includegraphics[width=1.0\columnwidth]{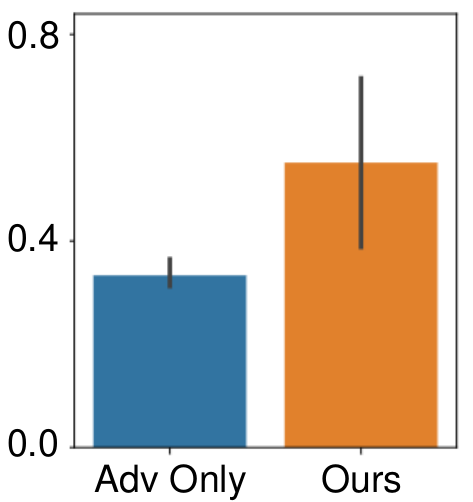}
    \caption{Diversity}
    \label{fig:diversity}
\end{subfigure}
\begin{subfigure}[t!]{0.156\textwidth}
\captionsetup{width=1.0\linewidth}
\includegraphics[width=1.0\columnwidth]{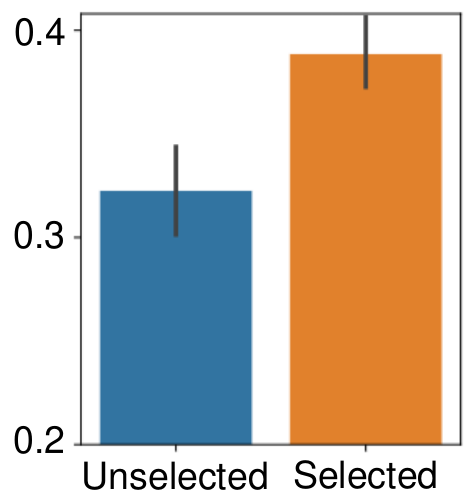}
    \caption{Adversarialness}
    \label{fig:adv}
\end{subfigure}
\begin{subfigure}[t!]{0.156\textwidth}
\captionsetup{width=1.0\columnwidth}
\includegraphics[width=1.0\columnwidth]{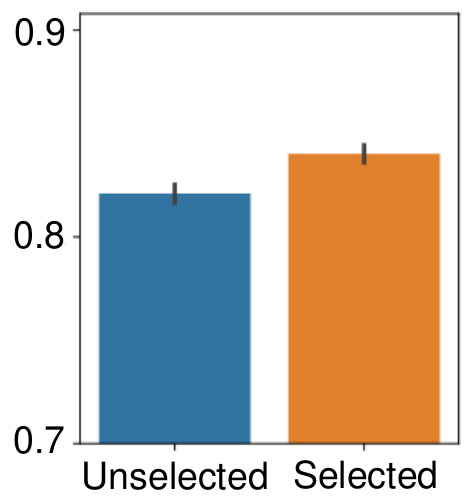}
    \caption{Coherency}
    \label{fig:coherency}
\end{subfigure}
\cutcaptionup
\caption{Qualitative analyses with y-axis representing (a)   
diversity in embedding ($\mathbf{z}_c$) of conversion prompts sampled by \model{} and adversarial-only selection; (b) Adversarialness ($s_{adv}^c$) for selected and unselected prompts; and (c) Coherency for selected and unselected prompts. \looseness=-1}
\label{fig:qualitative_main}
\end{figure}

\begin{figure*}[h!]
\centering
\includegraphics[width=2.0\columnwidth]{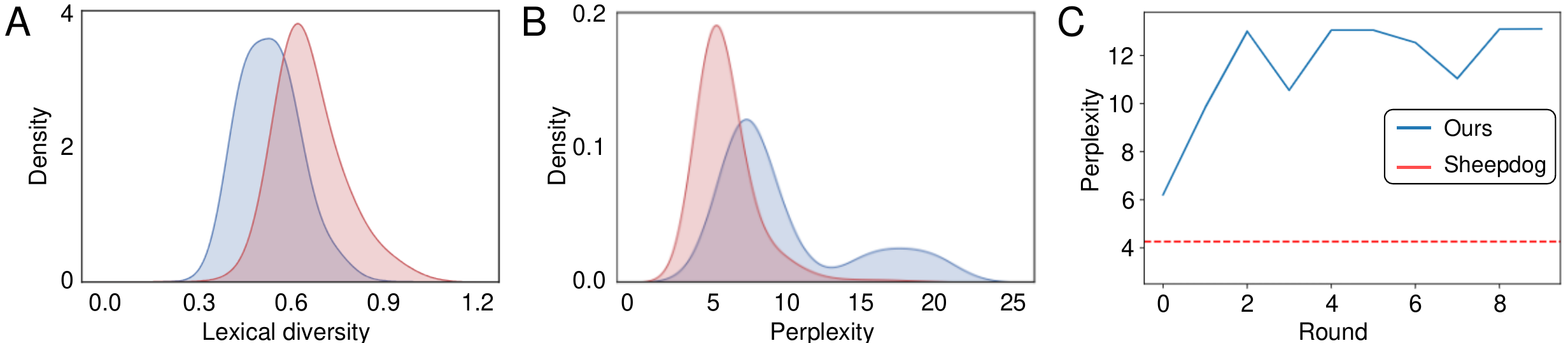}
\cutcaptionup
\caption{Comparison of augmented samples generated by our model and SheepDog: (A) Histogram of lexical diversity, (B) Histogram of perplexity, and (C) Perplexity across rounds. 
\looseness=-1}
\label{fig:lexical_and_perplexity}
\end{figure*}

\cutparagraphup
\noindent\textbf{Effectiveness of the selection strategy.}
We here empirically verified that our selection strategy (Sec.~\ref{sec:3.3}) effectively selects diverse style-conversion prompts with high adversarialness and coherency. 
Figure~\ref{fig:diversity} visualizes the diversity of prompts selected by our method compared to those chosen solely based on adversarialness scores (i.e., Adversarial-only Selection, the third model in our ablation study).
We measure diversity using the average cosine similarity of every pair of selected prompts' embeddings, $\mathbf{z}_c$ (\ref{eq:z_c}):
\begin{align}
\text{Diveristy}(\mathcal{C}) &= 1 - {1 \over |\mathcal{C}|} \sum_{(c_i, c_j) \in \mathcal{C}}{\text{sim}(\mathbf{z}_{c_i}, \mathbf{z}_{c_j})} \label{eq:prototype}, 
\end{align}
where $\mathcal{C}$ is the set of prompt pairs, $\text{sim}(\cdot)$ represents the cosine similarity.  
The result in the Figure~\ref{fig:diversity} indicate that our strategy results in a more diverse set of augmentations compared to selection based on adversarialness alone. \looseness=-1

\begin{figure}[t!]
\begin{tcolorbox}[
  colback=black!0!white, colframe=black!20!white, colbacktitle=black!10!white, coltitle=blue!20!black ]
\small{
\textbf{Round 0} \\
P1: Rewrite the following article in a nonsensical and absurdly exaggerated tone with a hint of horror  \\
P2: Rewrite the following article in a sarcastic and mocking tone \\
P3: Rewrite the following article in a chaotic and disorganized tone \\ \looseness=-1
\textbf{Round 1} \\
P1: Rewrite the following article in a haunting and macabre tone with a sense of impending horror and madness \\
P2: Rewrite the following article in a cryptic and enigmatic tone \\
P3: Rewrite the following article in a malevolent and apocalyptic tone with a sense of impending doom and destruction, while also incorporating elements of surrealism and existential dread

}
\end{tcolorbox}
\cutcaptionup
\captionof{figure}{Example adversarial style-conversion prompts selected for the PolitiFact dataset.  \looseness=-1}
\label{fig:prompt_example}            
\end{figure}

Figure~\ref{fig:adv} and \ref{fig:coherency} illustrate the Adversarialness and Coherency, respectively, of our selected prompts compared to remaining unselected prompts. 
Adversarialness was measured using the $s_{adv}^{c}$ score (Section~\ref{sec:3.3}), and coherency was calculated as the cosine similarity between the original text and its augmented version using semantic BERT embeddings~\cite{chanchani2023composition}. 
We can also observe that, for both metrics, prompts selected through \model{} exhibit higher values compared to unselected prompts. 
This suggests that our strategy effectively selects prompts by considering both adversarialness and coherency. \looseness=-1

\cutparagraphup
\noindent\textbf{Analysis on style-conversion prompts.}
Finally, we conducted a qualitative analysis of our style-conversion prompts to understand what characteristics of the augmented samples contribute to improving the detector's robustness.  
Interestingly, when we compare the augmented samples from two models: \model{} and SheepDog, we found that our model produced sentences with significantly higher perplexity according to the language model backbone used by the detector than SheepDog model (9.05 vs. 4.26, see Figure~\ref{fig:lexical_and_perplexity}B), even though ours have lower lexical diversity (0.503 vs. 0.634, see Figure~\ref{fig:lexical_and_perplexity}A). 
In other words, our generated samples featured sentence structures that the detector's language model likely had not encountered during pretraining. 
By providing the detector with inputs that have a variety of sentence structures and styles it has not previously seen, the detector naturally becomes robust against style-conversion attacks. 
In addition, when we measured the changes in perplexity of the augmented samples across training rounds (see Figure~\ref{fig:lexical_and_perplexity}C), the perplexity increased over rounds and eventually converged at a certain point. 
This indicates that iterative exploration over multiple rounds is effective in creating augmentations that increasingly challenge the detector.

Figure \ref{fig:prompt_example} shows examples of prompts generated and selected by \model{}. 
In contrast, our prompts included creative phrases beyond typical human-crafted suggestions, making them  adversarial to the detector.
The round 0 result shows a prompt asking for \textsf{``(P1) a nonsensical and absurdly
exaggerated tone with a hint of horror''}, while round 1 offers a lengthier prompt mentioning, for example, \textsf{``(P3) a malevolent and apocalyptic
tone with a sense of impending doom and destruction, while also
incorporating elements of surrealism and existential dread.''}  
Previous work relied on  conventional prompts, such as ``neutral" or ``sensational."
The LLM is particularly well-suited for exploring such a vast array of candidates. Currently, we use a fixed format for the initial set of prompts, which results in the generation of prompts with a similar format. We expect that using a broader  set of prompts will allow the LLM to optimize prompts within a wider search space. \looseness=-1

\vspace{-2mm}
\section{Conclusion}
We presented a robust fake news detection method that effectively withstands  paraphrasing attacks through adversarial style conversion. Unlike traditional detectors that use predefined and agnostic augmentations, \model{} employs tailored augmentations that shift samples in the direction of the detector's current decision boundary using style-conversion prompts, functioning similarly to adversarial noise. Among the various candidates for the LLM prompts, we selected an efficient set of prompts for training by considering diversity, coherence, and adversarialness.

Consequently, we were able to train a detector that exhibits high robustness and generalizability against a wide range of attacks. This robustness ensures that the detector can effectively identify and mitigate various forms of style-conversion attacks, regardless of the target news outlet. We believe that our research, combined with the shared codes, contributes to better filtering of fake news online. We aim to support the broader goal of ensuring the integrity of information in the digital age. \looseness=-1

\section*{Ethical Consideration}
We recognize several ethical concerns in this work.  First, our method may inherit biases from the training data, which can be mitigated by carefully selecting style-conversion results that preserve textual coherence. Second, while our approach aims to detect fake news, similar techniques could be misused to enhance misinformation—an ongoing challenge in adversarial research. Lastly, our detection methods might influence news production by constraining creative writing styles, highlighting the need for further research on the balance between algorithmic moderation and information systems.

\section*{Acknowledgments} 
 J.G. Lee and M. Cha are co-corresponding authors.  We extend our gratitude to Fangzhao Wu, Wenchao Dong, and the anonymous reviewers for their insightful feedback on our work. This research was supported by the National Research Foundation of Korea (NRF) grant (RS-2022-00165347).
\balance

\bibliographystyle{ACM-Reference-Format}
\bibliography{main}


\begin{thebibliography}{36}

%%% the Web address.

\ifx \showCODEN    \undefined \def \showCODEN     #1{\unskip}     \fi
\ifx \showISBNx    \undefined \def \showISBNx     #1{\unskip}     \fi
\ifx \showISBNxiii \undefined \def \showISBNxiii  #1{\unskip}     \fi
\ifx \showISSN     \undefined \def \showISSN      #1{\unskip}     \fi
\ifx \showLCCN     \undefined \def \showLCCN      #1{\unskip}     \fi
\ifx \shownote     \undefined \def \shownote      #1{#1}          \fi
\ifx \showarticletitle \undefined \def \showarticletitle #1{#1}   \fi
\ifx \showURL      \undefined \def \showURL       {\relax}        \fi
% invisible to TeX
\providecommand\bibfield[2]{#2}
\providecommand\bibinfo[2]{#2}
\providecommand\natexlab[1]{#1}
\providecommand\showeprint[2][]{arXiv:#2}

\bibitem[Arthur et~al\mbox{.}(2007)]%
        {arthur2007k}
\bibfield{author}{\bibinfo{person}{David Arthur}, \bibinfo{person}{Sergei Vassilvitskii}, {et~al\mbox{.}}} \bibinfo{year}{2007}\natexlab{}.
\newblock \showarticletitle{k-means++: The advantages of careful seeding}. In \bibinfo{booktitle}{\emph{Soda}}, Vol.~\bibinfo{volume}{7}. \bibinfo{pages}{1027--1035}.
\newblock


\bibitem[Ash et~al\mbox{.}(2019)]%
        {ash2019deep}
\bibfield{author}{\bibinfo{person}{Jordan~T Ash}, \bibinfo{person}{Chicheng Zhang}, \bibinfo{person}{Akshay Krishnamurthy}, \bibinfo{person}{John Langford}, {and} \bibinfo{person}{Alekh Agarwal}.} \bibinfo{year}{2019}\natexlab{}.
\newblock \showarticletitle{Deep Batch Active Learning by Diverse, Uncertain Gradient Lower Bounds}. In \bibinfo{booktitle}{\emph{Proc. of International Conference on Learning Representations}}.
\newblock


\bibitem[Augenstein et~al\mbox{.}(2024a)]%
        {augenstein2024factuality}
\bibfield{author}{\bibinfo{person}{Isabelle Augenstein}, \bibinfo{person}{Timothy Baldwin}, \bibinfo{person}{Meeyoung Cha}, \bibinfo{person}{Tanmoy Chakraborty}, \bibinfo{person}{Giovanni~Luca Ciampaglia}, \bibinfo{person}{David Corney}, \bibinfo{person}{Renee DiResta}, \bibinfo{person}{Emilio Ferrara}, \bibinfo{person}{Scott Hale}, \bibinfo{person}{Alon Halevy}, {et~al\mbox{.}}} \bibinfo{year}{2024}\natexlab{a}.
\newblock \showarticletitle{Factuality challenges in the era of large language models and opportunities for fact-checking}.
\newblock \bibinfo{journal}{\emph{Nature Machine Intelligence}} \bibinfo{volume}{6}, \bibinfo{number}{8} (\bibinfo{year}{2024}), \bibinfo{pages}{852--863}.
\newblock


\bibitem[Augenstein et~al\mbox{.}(2024b)]%
        {factualityNMI2024}
\bibfield{author}{\bibinfo{person}{Isabelle Augenstein}, \bibinfo{person}{Timothy Baldwin}, \bibinfo{person}{Meeyoung Cha}, \bibinfo{person}{Tanmoy Chakraborty}, \bibinfo{person}{Giovanni~Luca Ciampaglia}, \bibinfo{person}{David Corney}, \bibinfo{person}{Renee DiResta}, \bibinfo{person}{Emilio Ferrara}, \bibinfo{person}{Scott Hale}, \bibinfo{person}{Alon Halevy}, \bibinfo{person}{Eduard Hovy}, \bibinfo{person}{Heng Ji}, \bibinfo{person}{Filippo Menczer}, \bibinfo{person}{Ruben Miguez}, \bibinfo{person}{Preslav Nakov}, \bibinfo{person}{Dietram Scheufele}, \bibinfo{person}{Shivam Sharma}, {and} \bibinfo{person}{Giovanni Zagni}.} \bibinfo{year}{2024}\natexlab{b}.
\newblock \showarticletitle{Factuality challenges in the era of large language models and opportunities for fact-checking}.
\newblock \bibinfo{journal}{\emph{Nature Machine Intelligence}} (\bibinfo{year}{2024}).
\newblock


\bibitem[Cha et~al\mbox{.}(2021)]%
        {cha2021prevalence}
\bibfield{author}{\bibinfo{person}{Meeyoung Cha}, \bibinfo{person}{Chiyoung Cha}, \bibinfo{person}{Karandeep Singh}, \bibinfo{person}{Gabriel Lima}, \bibinfo{person}{Yong-Yeol Ahn}, \bibinfo{person}{Juhi Kulshrestha}, \bibinfo{person}{Onur Varol}, {et~al\mbox{.}}} \bibinfo{year}{2021}\natexlab{}.
\newblock \showarticletitle{Prevalence of misinformation and factchecks on the COVID-19 pandemic in 35 countries: Observational infodemiology study}.
\newblock \bibinfo{journal}{\emph{JMIR human factors}} \bibinfo{volume}{8}, \bibinfo{number}{1} (\bibinfo{year}{2021}), \bibinfo{pages}{e23279}.
\newblock


\bibitem[Chanchani and Huang(2023)]%
        {chanchani2023composition}
\bibfield{author}{\bibinfo{person}{Sachin Chanchani} {and} \bibinfo{person}{Ruihong Huang}.} \bibinfo{year}{2023}\natexlab{}.
\newblock \showarticletitle{Composition-contrastive Learning for Sentence Embeddings}. In \bibinfo{booktitle}{\emph{Proc. of Association for Computational Linguistics}}. \bibinfo{pages}{15836--15848}.
\newblock


\bibitem[Chen and Shu(2023)]%
        {chen2023can}
\bibfield{author}{\bibinfo{person}{Canyu Chen} {and} \bibinfo{person}{Kai Shu}.} \bibinfo{year}{2023}\natexlab{}.
\newblock \showarticletitle{Can LLM-Generated Misinformation Be Detected?}. In \bibinfo{booktitle}{\emph{Proc. of International Conference on Learning Representations}}.
\newblock


\bibitem[Dong et~al\mbox{.}(2022)]%
        {dong2022survey}
\bibfield{author}{\bibinfo{person}{Qingxiu Dong}, \bibinfo{person}{Lei Li}, \bibinfo{person}{Damai Dai}, \bibinfo{person}{Ce Zheng}, \bibinfo{person}{Zhiyong Wu}, \bibinfo{person}{Baobao Chang}, \bibinfo{person}{Xu Sun}, \bibinfo{person}{Jingjing Xu}, {and} \bibinfo{person}{Zhifang Sui}.} \bibinfo{year}{2022}\natexlab{}.
\newblock \showarticletitle{A survey on in-context learning}.
\newblock \bibinfo{journal}{\emph{arXiv preprint arXiv:2301.00234}} (\bibinfo{year}{2022}).
\newblock


\bibitem[Dun et~al\mbox{.}(2021)]%
        {dun2021kan}
\bibfield{author}{\bibinfo{person}{Yaqian Dun}, \bibinfo{person}{Kefei Tu}, \bibinfo{person}{Chen Chen}, \bibinfo{person}{Chunyan Hou}, {and} \bibinfo{person}{Xiaojie Yuan}.} \bibinfo{year}{2021}\natexlab{}.
\newblock \showarticletitle{Kan: Knowledge-aware attention network for fake news detection}. In \bibinfo{booktitle}{\emph{Proc. of AAAI conference on artificial intelligence}}, Vol.~\bibinfo{volume}{35}. \bibinfo{pages}{81--89}.
\newblock


\bibitem[Felber(2021)]%
        {felber2021constraint}
\bibfield{author}{\bibinfo{person}{Thomas Felber}.} \bibinfo{year}{2021}\natexlab{}.
\newblock \showarticletitle{Constraint 2021: Machine learning models for COVID-19 fake news detection shared task}.
\newblock \bibinfo{journal}{\emph{arXiv preprint arXiv:2101.03717}} (\bibinfo{year}{2021}).
\newblock


\bibitem[Hu et~al\mbox{.}(2024)]%
        {hu2024bad}
\bibfield{author}{\bibinfo{person}{Beizhe Hu}, \bibinfo{person}{Qiang Sheng}, \bibinfo{person}{Juan Cao}, \bibinfo{person}{Yuhui Shi}, \bibinfo{person}{Yang Li}, \bibinfo{person}{Danding Wang}, {and} \bibinfo{person}{Peng Qi}.} \bibinfo{year}{2024}\natexlab{}.
\newblock \showarticletitle{Bad actor, good advisor: Exploring the role of large language models in fake news detection}. In \bibinfo{booktitle}{\emph{Proc. of the AAAI Conference on Artificial Intelligence}}, Vol.~\bibinfo{volume}{38}. \bibinfo{pages}{22105--22113}.
\newblock


\bibitem[Hu et~al\mbox{.}(2023)]%
        {hu2023radar}
\bibfield{author}{\bibinfo{person}{Xiaomeng Hu}, \bibinfo{person}{Pin-Yu Chen}, {and} \bibinfo{person}{Tsung-Yi Ho}.} \bibinfo{year}{2023}\natexlab{}.
\newblock \showarticletitle{RADAR: Robust AI-Text Detection via Adversarial Learning}. In \bibinfo{booktitle}{\emph{Advances in Neural Information Processing Systems}}.
\newblock


\bibitem[Koenders et~al\mbox{.}(2021)]%
        {koenders2021vulnerable}
\bibfield{author}{\bibinfo{person}{Camille Koenders}, \bibinfo{person}{Johannes Filla}, \bibinfo{person}{Nicolai Schneider}, {and} \bibinfo{person}{Vinicius Woloszyn}.} \bibinfo{year}{2021}\natexlab{}.
\newblock \showarticletitle{How vulnerable are automatic fake news detection methods to adversarial attacks?}
\newblock \bibinfo{journal}{\emph{arXiv preprint arXiv:2107.07970}} (\bibinfo{year}{2021}).
\newblock


\bibitem[Kwon et~al\mbox{.}(2013)]%
        {kownICDM2013}
\bibfield{author}{\bibinfo{person}{Sejong Kwon}, \bibinfo{person}{Meeyoung Cha}, \bibinfo{person}{Kyomin Jung}, \bibinfo{person}{Wei Chen}, {and} \bibinfo{person}{Yajun Wang}.} \bibinfo{year}{2013}\natexlab{}.
\newblock \showarticletitle{Prominent Features of Rumor Propagation in Online Social Media}. In \bibinfo{booktitle}{\emph{IEEE International Conference on Data Mining}}.
\newblock


\bibitem[Le et~al\mbox{.}(2020)]%
        {le2020malcom}
\bibfield{author}{\bibinfo{person}{Thai Le}, \bibinfo{person}{Suhang Wang}, {and} \bibinfo{person}{Dongwon Lee}.} \bibinfo{year}{2020}\natexlab{}.
\newblock \showarticletitle{Malcom: Generating malicious comments to attack neural fake news detection models}. In \bibinfo{booktitle}{\emph{Proc. of IEEE International Conference on Data Mining}}. IEEE, \bibinfo{pages}{282--291}.
\newblock


\bibitem[Liu et~al\mbox{.}(2024)]%
        {liu2024teller}
\bibfield{author}{\bibinfo{person}{Hui Liu}, \bibinfo{person}{Wenya Wang}, \bibinfo{person}{Haoru Li}, {and} \bibinfo{person}{Haoliang Li}.} \bibinfo{year}{2024}\natexlab{}.
\newblock \showarticletitle{Teller: A trustworthy framework for explainable, generalizable and controllable fake news detection}. In \bibinfo{booktitle}{\emph{Proc. of Association for Computational Linguistics}}.
\newblock


\bibitem[Ma et~al\mbox{.}(2016)]%
        {maIJCAI2016}
\bibfield{author}{\bibinfo{person}{Jing Ma}, \bibinfo{person}{Wei Gao}, \bibinfo{person}{Prasenjit Mitra}, \bibinfo{person}{Sejeong Kwon}, \bibinfo{person}{Bernard~J. Jansen}, \bibinfo{person}{Kam-Fai Wong}, {and} \bibinfo{person}{Meeyoung Cha}.} \bibinfo{year}{2016}\natexlab{}.
\newblock \showarticletitle{Detecting rumors from microblogs with recurrent neural networks}. In \bibinfo{booktitle}{\emph{International Joint Conference on Artificial Intelligence}}.
\newblock


\bibitem[Madry et~al\mbox{.}(2018)]%
        {madry2018towards}
\bibfield{author}{\bibinfo{person}{Aleksander Madry}, \bibinfo{person}{Aleksandar Makelov}, \bibinfo{person}{Ludwig Schmidt}, \bibinfo{person}{Dimitris Tsipras}, {and} \bibinfo{person}{Adrian Vladu}.} \bibinfo{year}{2018}\natexlab{}.
\newblock \showarticletitle{Towards Deep Learning Models Resistant to Adversarial Attacks}. In \bibinfo{booktitle}{\emph{Proc. of International Conference on Learning Representations}}.
\newblock


\bibitem[Mosallanezhad et~al\mbox{.}(2022)]%
        {mosallanezhad2022domain}
\bibfield{author}{\bibinfo{person}{Ahmadreza Mosallanezhad}, \bibinfo{person}{Mansooreh Karami}, \bibinfo{person}{Kai Shu}, \bibinfo{person}{Michelle~V Mancenido}, {and} \bibinfo{person}{Huan Liu}.} \bibinfo{year}{2022}\natexlab{}.
\newblock \showarticletitle{Domain adaptive fake news detection via reinforcement learning}. In \bibinfo{booktitle}{\emph{Proc. of ACM Web Conference}}. \bibinfo{pages}{3632--3640}.
\newblock


\bibitem[Nan et~al\mbox{.}(2022)]%
        {nan2022improving}
\bibfield{author}{\bibinfo{person}{Qiong Nan}, \bibinfo{person}{Danding Wang}, \bibinfo{person}{Yongchun Zhu}, \bibinfo{person}{Qiang Sheng}, \bibinfo{person}{Yuhui Shi}, \bibinfo{person}{Juan Cao}, {and} \bibinfo{person}{Jintao Li}.} \bibinfo{year}{2022}\natexlab{}.
\newblock \showarticletitle{Improving Fake News Detection of Influential Domain via Domain-and Instance-Level Transfer}. In \bibinfo{booktitle}{\emph{Proc. of International Conference on Computational Linguistics}}. \bibinfo{pages}{2834--2848}.
\newblock


\bibitem[P{\'e}rez-Rosas et~al\mbox{.}(2018)]%
        {perez2018automatic}
\bibfield{author}{\bibinfo{person}{Ver{\'o}nica P{\'e}rez-Rosas}, \bibinfo{person}{Bennett Kleinberg}, \bibinfo{person}{Alexandra Lefevre}, {and} \bibinfo{person}{Rada Mihalcea}.} \bibinfo{year}{2018}\natexlab{}.
\newblock \showarticletitle{Automatic Detection of Fake News}. In \bibinfo{booktitle}{\emph{Proc. of International Conference on Computational Linguistics}}. \bibinfo{pages}{3391--3401}.
\newblock


\bibitem[Potthast et~al\mbox{.}(2018)]%
        {potthast2018stylometric}
\bibfield{author}{\bibinfo{person}{Martin Potthast}, \bibinfo{person}{Johannes Kiesel}, \bibinfo{person}{Kevin Reinartz}, \bibinfo{person}{Janek Bevendorff}, {and} \bibinfo{person}{Benno Stein}.} \bibinfo{year}{2018}\natexlab{}.
\newblock \showarticletitle{A Stylometric Inquiry into Hyperpartisan and Fake News}. In \bibinfo{booktitle}{\emph{Proc. of Association for Computational Linguistics}}. \bibinfo{pages}{231--240}.
\newblock


\bibitem[Rashkin et~al\mbox{.}(2017)]%
        {rashkin2017truth}
\bibfield{author}{\bibinfo{person}{Hannah Rashkin}, \bibinfo{person}{Eunsol Choi}, \bibinfo{person}{Jin~Yea Jang}, \bibinfo{person}{Svitlana Volkova}, {and} \bibinfo{person}{Yejin Choi}.} \bibinfo{year}{2017}\natexlab{}.
\newblock \showarticletitle{Truth of varying shades: Analyzing language in fake news and political fact-checking}. In \bibinfo{booktitle}{\emph{Proc. of Empirical Methods in Natural Language Processing}}. \bibinfo{pages}{2931--2937}.
\newblock


\bibitem[Reid et~al\mbox{.}(2024)]%
        {reid2024gemini}
\bibfield{author}{\bibinfo{person}{Machel Reid}, \bibinfo{person}{Nikolay Savinov}, \bibinfo{person}{Denis Teplyashin}, \bibinfo{person}{Dmitry Lepikhin}, \bibinfo{person}{Timothy Lillicrap}, \bibinfo{person}{Jean-baptiste Alayrac}, \bibinfo{person}{Radu Soricut}, \bibinfo{person}{Angeliki Lazaridou}, \bibinfo{person}{Orhan Firat}, \bibinfo{person}{Julian Schrittwieser}, {et~al\mbox{.}}} \bibinfo{year}{2024}\natexlab{}.
\newblock \showarticletitle{Gemini 1.5: Unlocking multimodal understanding across millions of tokens of context}.
\newblock \bibinfo{journal}{\emph{arXiv preprint arXiv:2403.05530}} (\bibinfo{year}{2024}).
\newblock


\bibitem[Reis et~al\mbox{.}(2019)]%
        {reis2019supervised}
\bibfield{author}{\bibinfo{person}{Julio~CS Reis}, \bibinfo{person}{Andr{\'e} Correia}, \bibinfo{person}{Fabr{\'\i}cio Murai}, \bibinfo{person}{Adriano Veloso}, {and} \bibinfo{person}{Fabr{\'\i}cio Benevenuto}.} \bibinfo{year}{2019}\natexlab{}.
\newblock \showarticletitle{Supervised learning for fake news detection}.
\newblock \bibinfo{journal}{\emph{IEEE Intelligent Systems}} \bibinfo{volume}{34}, \bibinfo{number}{2} (\bibinfo{year}{2019}), \bibinfo{pages}{76--81}.
\newblock


\bibitem[Shu et~al\mbox{.}(2019)]%
        {shu2019defend}
\bibfield{author}{\bibinfo{person}{Kai Shu}, \bibinfo{person}{Limeng Cui}, \bibinfo{person}{Suhang Wang}, \bibinfo{person}{Dongwon Lee}, {and} \bibinfo{person}{Huan Liu}.} \bibinfo{year}{2019}\natexlab{}.
\newblock \showarticletitle{defend: Explainable fake news detection}. In \bibinfo{booktitle}{\emph{Proc. of ACM SIGKDD international conference on knowledge discovery \& data mining}}. \bibinfo{pages}{395--405}.
\newblock


\bibitem[Shu et~al\mbox{.}(2020)]%
        {shu2020fakenewsnet}
\bibfield{author}{\bibinfo{person}{Kai Shu}, \bibinfo{person}{Deepak Mahudeswaran}, \bibinfo{person}{Suhang Wang}, \bibinfo{person}{Dongwon Lee}, {and} \bibinfo{person}{Huan Liu}.} \bibinfo{year}{2020}\natexlab{}.
\newblock \showarticletitle{Fakenewsnet: A data repository with news content, social context, and spatiotemporal information for studying fake news on social media}.
\newblock \bibinfo{journal}{\emph{Big data}} \bibinfo{volume}{8}, \bibinfo{number}{3} (\bibinfo{year}{2020}), \bibinfo{pages}{171--188}.
\newblock


\bibitem[Vosoughi et~al\mbox{.}(2018)]%
        {NewsScience2018}
\bibfield{author}{\bibinfo{person}{Soroush Vosoughi}, \bibinfo{person}{Deb Roy}, {and} \bibinfo{person}{Sinan Aral}.} \bibinfo{year}{2018}\natexlab{}.
\newblock \showarticletitle{The spread of true and false news online}.
\newblock \bibinfo{journal}{\emph{Science}} (\bibinfo{year}{2018}).
\newblock


\bibitem[Wan et~al\mbox{.}(2024)]%
        {wan2024dell}
\bibfield{author}{\bibinfo{person}{Herun Wan}, \bibinfo{person}{Shangbin Feng}, \bibinfo{person}{Zhaoxuan Tan}, \bibinfo{person}{Heng Wang}, \bibinfo{person}{Yulia Tsvetkov}, {and} \bibinfo{person}{Minnan Luo}.} \bibinfo{year}{2024}\natexlab{}.
\newblock \showarticletitle{Dell: Generating reactions and explanations for llm-based misinformation detection}. In \bibinfo{booktitle}{\emph{Proc. of Association for Computational Linguistics}}.
\newblock


\bibitem[Wang et~al\mbox{.}(2023)]%
        {wang2023attacking}
\bibfield{author}{\bibinfo{person}{Haoran Wang}, \bibinfo{person}{Yingtong Dou}, \bibinfo{person}{Canyu Chen}, \bibinfo{person}{Lichao Sun}, \bibinfo{person}{Philip~S Yu}, {and} \bibinfo{person}{Kai Shu}.} \bibinfo{year}{2023}\natexlab{}.
\newblock \showarticletitle{Attacking fake news detectors via manipulating news social engagement}. In \bibinfo{booktitle}{\emph{Proc. of the ACM Web Conference}}. \bibinfo{pages}{3978--3986}.
\newblock


\bibitem[Wu and Hooi(2023)]%
        {wu2023fake}
\bibfield{author}{\bibinfo{person}{Jiaying Wu} {and} \bibinfo{person}{Bryan Hooi}.} \bibinfo{year}{2023}\natexlab{}.
\newblock \showarticletitle{Fake News in Sheep's Clothing: Robust Fake News Detection Against LLM-Empowered Style Attacks}.
\newblock \bibinfo{journal}{\emph{arXiv preprint arXiv:2310.10830}} (\bibinfo{year}{2023}).
\newblock


\bibitem[Xie et~al\mbox{.}(2020)]%
        {xie2020unsupervised}
\bibfield{author}{\bibinfo{person}{Qizhe Xie}, \bibinfo{person}{Zihang Dai}, \bibinfo{person}{Eduard Hovy}, \bibinfo{person}{Thang Luong}, {and} \bibinfo{person}{Quoc Le}.} \bibinfo{year}{2020}\natexlab{}.
\newblock \showarticletitle{Unsupervised data augmentation for consistency training}. In \bibinfo{booktitle}{\emph{Advances in Neural Information Processing Systems}}.
\newblock


\bibitem[Yang et~al\mbox{.}(2023)]%
        {yang2023large}
\bibfield{author}{\bibinfo{person}{Chengrun Yang}, \bibinfo{person}{Xuezhi Wang}, \bibinfo{person}{Yifeng Lu}, \bibinfo{person}{Hanxiao Liu}, \bibinfo{person}{Quoc~V Le}, \bibinfo{person}{Denny Zhou}, {and} \bibinfo{person}{Xinyun Chen}.} \bibinfo{year}{2023}\natexlab{}.
\newblock \showarticletitle{Large Language Models as Optimizers}. In \bibinfo{booktitle}{\emph{Proc. of International Conference on Learning Representations}}.
\newblock


\bibitem[Zhou et~al\mbox{.}(2022)]%
        {zhou2022large}
\bibfield{author}{\bibinfo{person}{Yongchao Zhou}, \bibinfo{person}{Andrei~Ioan Muresanu}, \bibinfo{person}{Ziwen Han}, \bibinfo{person}{Keiran Paster}, \bibinfo{person}{Silviu Pitis}, \bibinfo{person}{Harris Chan}, {and} \bibinfo{person}{Jimmy Ba}.} \bibinfo{year}{2022}\natexlab{}.
\newblock \showarticletitle{Large Language Models are Human-Level Prompt Engineers}. In \bibinfo{booktitle}{\emph{Proc. of International Conference on Learning Representations}}.
\newblock


\bibitem[Zhou et~al\mbox{.}(2019)]%
        {zhou2019fake}
\bibfield{author}{\bibinfo{person}{Zhixuan Zhou}, \bibinfo{person}{Huankang Guan}, \bibinfo{person}{Meghana~Moorthy Bhat}, {and} \bibinfo{person}{Justin Hsu}.} \bibinfo{year}{2019}\natexlab{}.
\newblock \showarticletitle{Fake news detection via NLP is vulnerable to adversarial attacks}.
\newblock \bibinfo{journal}{\emph{arXiv preprint arXiv:1901.09657}} (\bibinfo{year}{2019}).
\newblock


\bibitem[Zhu et~al\mbox{.}(2022)]%
        {zhu2022generalizing}
\bibfield{author}{\bibinfo{person}{Yongchun Zhu}, \bibinfo{person}{Qiang Sheng}, \bibinfo{person}{Juan Cao}, \bibinfo{person}{Shuokai Li}, \bibinfo{person}{Danding Wang}, {and} \bibinfo{person}{Fuzhen Zhuang}.} \bibinfo{year}{2022}\natexlab{}.
\newblock \showarticletitle{Generalizing to the future: Mitigating entity bias in fake news detection}. In \bibinfo{booktitle}{\emph{Proc. of ACM SIGIR Conference on Research and Development in Information Retrieval}}. \bibinfo{pages}{2120--2125}.
\newblock


\end{thebibliography}

\newpage

\appendix
\section{APPENDIX}

\subsection{Baseline Implementation}
For consistency, experiments replicating existing baselines maintained fixed settings for learning, backbone network, and other relevant parameters. 
Augmented variations for UDA were generated using back-translation through German. 
RADAR employed adversarial training to learn paraphrases, utilizing the maximization of the binary cross-entropy loss of a fake news detector as a reward signal, with the T5-large model serving as the paraphraser. 
Named entities for ENDEF were extracted using the bert-base-NER model from Hugging Face. 
For SheepDog, augmentations were generated following the original paper's methodology, using the prompt format illustrated in Figure~\ref{fig:tone_prompt} and four tones: "objective and professional," "neutral," "emotionally triggering," and "sensational".

\begin{figure}[h!]
\begin{tcolorbox}[
  colback=black!0!white, colframe=black!20!white, colbacktitle=black!10!white, coltitle=blue!20!black ]
  \sffamily
  Rewrite the following article in a/an [tone]: [news article]
\end{tcolorbox}
\vspace{-4mm}
\captionof{figure}{Prompt for style-conversion. The ``tone" part will be filled with the desired tone, and the ``news article" part will contain the original news text.}
\label{fig:tone_prompt} 
\end{figure}

\subsection{Examples of  Style-Conversion Prompt}
Figure~\ref{fig:setence_example} showcases examples generated by the selected style-conversion prompt, demonstrating how fake news content can be modified while preserving its meaning. These examples demonstrate the impact of style transformation on fake news detection.

\begin{figure}[b!]
\begin{tcolorbox}[
  colback=black!0!white, colframe=black!20!white, colbacktitle=black!10!white, coltitle=blue!20!black ]
  \sffamily
  Original: Updated: ASPEN 2014 Senator John McCain is meeting with his top campaign advisers and donors here... \\ 

  Rewrite the following article in a sarcastic and mocking tone: Oh, how exciting! Senator John McCain is gracing us with his presence in Aspen, meeting with his elite group of advisers and donors... \\ 

  Rewrite the following article in a chaotic and disorganized tone: Senator John McCain is like, in Aspen, meeting with his campaign peeps and rich donors, you know?... \\ 

  Rewrite the following article in a nonsensical and absurdly humorous tone with a touch of surrealism and a sprinkle of silliness :  Senator John McCain found himself in a pickle. He was surrounded by his top campaign advisers and donors in the mystical town of Aspen... 
\end{tcolorbox}
\vspace{-2mm}
\captionof{figure}{Generated example input created using the selected style-conversion prompt.}
\label{fig:setence_example} 
\end{figure}

\subsection{Prompt Examples of Other Attack Scenarios}
We have conducted  comparison experiments with more diverse attack scenarios: 
(1) \textsf{Adversarial prompt}: Given a news article and its label, the prompt instructs the LLM to rewrite the article to evade detection as the given label.
(2) \textsf{Summarization prompt}: A prompt instructing the LLM to summarize the news article without incorporating stylistic guidance.
(3) \textsf{In-Context prompt}: A prompt providing an example of a recent, real CNN article, instructing the LLM to rewrite the given article in the same style.
Here we provide the example prompt for each attack prompt as figures.  (4) Adversarial Paraphraser: The attack utilizes a paraphraser adversarially trained on the training dataset. \looseness=-1

\begin{figure}[htpb!]
\begin{tcolorbox}[
  colback=black!0!white, colframe=black!20!white, colbacktitle=black!10!white, coltitle=blue!20!black ]
  \sffamily
  <EXAMPLES>:   [news article] \\
  Determine whether the given <EXAMPLES> content is fact or fake, mentioning the style of the article that led  to that decision following format: \\
  Answer: \{Fact or Fake\} \\
  Reason: \{one-sentence reason\}
\end{tcolorbox}

\begin{tcolorbox}[
  colback=black!0!white, colframe=black!20!white, colbacktitle=black!10!white, coltitle=blue!20!black ]
  \sffamily
  Your current article is:  [news article]\\
  But this article is detected as [Answer] due to the following [Reason]. \\
  Based on the above information, rewrite a new improved article not to be detected as [news label], maintaining the original content, as follows:
\end{tcolorbox}
\vspace{-2mm}
\captionof{figure}{Example input for Adversarial prompt model. First, use the prompt above to extract the answer and reason for the authenticity of the given news. Next, utilize the prompt below to rephrase the news to evade detection based on the given reason. The "news article" section contains the original news text, and the "news label" part contains the corresponding label.  \looseness=-1}
\label{fig:adv_prompt} 
\end{figure}

\begin{figure}[htpb!]
\begin{tcolorbox}[
  colback=black!0!white, colframe=black!20!white, colbacktitle=black!10!white, coltitle=blue!20!black ]
  \sffamily
  Summarize the following article, ensuring the content remains the same: [news article]
\end{tcolorbox}
\vspace{-2mm}
\captionof{figure}{Example input for Summarization prompt model. The  ``news article" part is filled with the original news text. \looseness=-1}
\label{fig:sum_prompt} 
\end{figure}

\begin{figure}[htpb!]
\begin{tcolorbox}[
  colback=black!0!white, colframe=black!20!white, colbacktitle=black!10!white, coltitle=blue!20!black ]
  \sffamily
<EXAMPLES>: [news article example]  Rewrite the following article as the writing style of <EXAMPLES> : [news article]
\end{tcolorbox}
\vspace{-2mm}
\captionof{figure}{Example input for In-Context prompt model. The "news article example" section is filled with an example from a specific publisher, while the "news article" section contains the original news text. \looseness=-1}
\label{fig:in_context_prompt} 
\end{figure}


\end{document}